\documentclass[letterpaper, 10 pt, conference]{ieeeconf}  

\IEEEoverridecommandlockouts                              

\usepackage{graphicx}
\usepackage{amsmath} 
\usepackage{amssymb}  
\usepackage{url}

\title{\LARGE \bf
Translating Natural Language Instructions to Computer Programs for Robot Manipulation
}

\author{Sagar Gubbi Venkatesh$^{1,2}$, Raviteja Upadrashta$^{2}$, and Bharadwaj Amrutur$^{1,2}$
\thanks{$^{1}$Department of Electrical and Communication Engineering, Indian Institute of Science, Bangalore 560012, India
        {\tt\small sagar@iisc.ac.in};}%
\thanks{$^{2}$Robert Bosch Center for Cyber Physical Systems, Indian Institute of Science, Bangalore 560012, India
         {\tt\small ravitejaupadras@iisc.ac.in}; {\tt\small amrutur@iisc.ac.in}}%

\thanks{**This work was supported by Robert Bosch Center for Cyber-Physical Systems}
}

\begin{document}

\maketitle
\thispagestyle{empty}
\pagestyle{empty}

\begin{abstract}

It is highly desirable for robots that work alongside humans to be able to understand instructions in natural language. Existing language conditioned imitation learning models directly predict the actuator commands from the image observation and the instruction text. Rather than directly predicting actuator commands, we propose translating the natural language instruction to a Python function which queries the scene by accessing the output of the object detector and controls the robot to perform the specified task. This enables the use of non-differentiable modules such as a constraint solver when computing commands to the robot. Moreover, the labels in this setup are significantly more informative computer programs that capture the intent of the expert rather than teleoperated demonstrations. We show that the proposed method performs better than training a neural network to directly predict the robot actions.


\end{abstract}

\section{INTRODUCTION}

A robot that can operate alongside humans and perform a variety of tasks in unconstrained environments is a long standing vision of robotic learning. These robots need to be capable of understanding instructions in natural language from untrained users\cite{lynch2020grounding}. In this paper, we address the problem of programming robots using natural language.

Imitation learning has been used in recent years to learn end-to-end visuomotor policies that directly map pixels to robot actuator commands\cite{levine2016end}\cite{zhang2018deep}\cite{yu2018one}\cite{rahmatizadeh2018virtual}\cite{giusti2015machine}\cite{venkatesh2020one}. However, this is not the only way neural networks can be used for controlling robots. It is also possible to use sensor data such as the camera feed to construct a vector space representation of the world and then to plan a path in this space\cite{bejjani2019}. For example, an object detector can be used to find all the objects in the scene. The locations of the detected objects are used to determine the robot motion necessary to move the objects to particular positions. Although this introduces rigidity in the representation of the world, the advantages of this approach include modularity (the object detector can be replaced without modifying the rest of the system) and interpretability (the output of the object detector can be examined separately).

\begin{figure}[!t]
    \centering
    \includegraphics[width=0.7\linewidth]{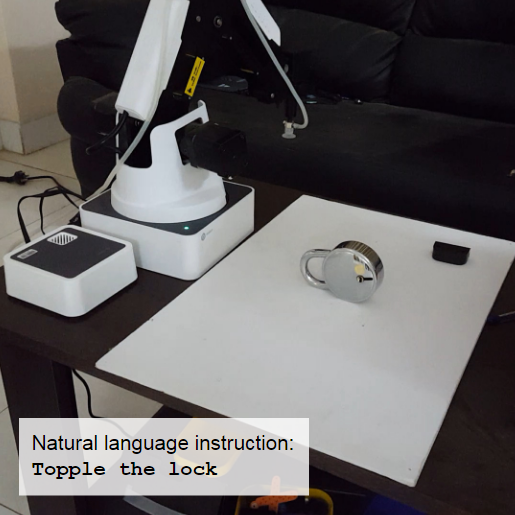}
    \caption{The robot receives an instruction in natural language, say ``Topple the lock", observes the scene through the camera, localizes the objects in front of it, translates the natural language instruction to a Python function block, and then executes the function.}
    \label{fig:hero}
\end{figure}

\begin{figure}[!t]
    \centering
    \includegraphics[width=1.0\linewidth]{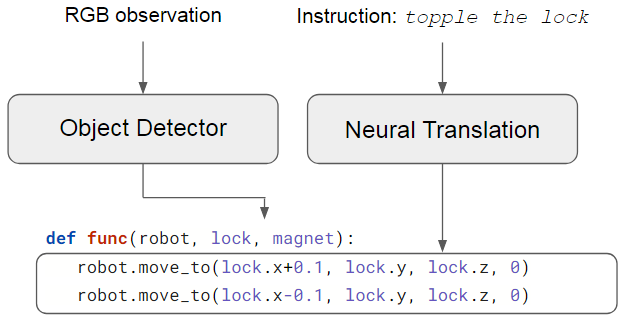}
    \caption{The instruction in natural language is translated into a Python function block, and the output of the object detector is passed as arguments to the function.}
    \label{fig:hero2}
\end{figure}

The majority of recent works on imitation learning have used some input device such as game controller\cite{rahmatizadeh2018vision}, VR controller\cite{zhang2018deep}, visual odometry based 6-DoF position tracking using smartphones\cite{mandlekar2018roboturk}\cite{mandlekar2019scaling}, space mouse\cite{gubbi2020imitation}, etc. to record experts teleoperating a robot. In this work, we take a different approach to collect expert demonstrations. We give a natural language instruction prompt and have experts write a Python function that controls the robot to accomplish the task specified in the instruction (Fig.~\ref{fig:hero}). This function takes the output of an object detector as its argument and moves the end-effector of the robot arm to perform the specified task (Fig.~\ref{fig:hero2}). The dataset collected in this manner is used to train a neural network that takes a natural language instruction as input and predicts a Python function block which controls the robot when executed.

A few examples of the tasks we consider are: (a) Push the orange towards the apple, (b) Place the apple between the orange and the apple, (c) Pick up the orange and use it to push the bottle off the edge of the table. Although our robot does not use a force sensor and can only move the end-effector using position control, it is possible to expand the set of primitive instructions of the robot to include complex macro instructions such as peg-in-hole insert instruction that may invoke a separately trained policy network\cite{gubbi2020imitation}. Our approach is most suitable for ``gluing" together simpler commands to compose a more complex program. A potential application for our method is in augmenting teach pendants to accept instructions in natural language.

There are several advantages of having expert demonstrations in the form of program code. One is that the expert program can invoke complex subroutines such as a constraint solver. It can be difficult to train an end-to-end neural network to copy the behavior of such complex modules. The other advantage is that the intention of the expert is clearer and less ambiguous in the program representation than in teleoperated demonstrations. For example, to ``push the orange off the table", the program to perform this task clearly indicates the robot motion for different possible positions of the orange, whereas, we would need many more teleoperated demonstrations each corresponding to a different position of the object to be able to train a neural network to reliably copy the expert behavior. Finally, the program representation is more interpretable and amenable to analysis before it is executed.

Our contributions are:
\begin{itemize}
    \item We propose an imitation learning setup where the expert demonstrations are in the form of program code and use a neural translation model to translate instructions in English to Python code that controls the robot.
    \item We show that the proposed method performs better than directly mapping natural language instructions to actuation commands.
\end{itemize}

The rest of this paper is organised as follows. In the following section, related work is discussed. Section~3 defines the problem statement. In Section~4, the neural network architecture that we use is described in detail. Experimental results are discussed in Section~5, and Section~6 concludes the paper.







\section{RELATED WORK}

Several recent papers have demonstrated that it is possible to learn visuomotor skills from human demonstrations\cite{giusti2015machine}\cite{bojarski2016end}\cite{daml}\cite{gubbi2020imitation}\cite{rahmatizadeh2018virtual}\cite{rahmatizadeh2018vision}\cite{venkatesh2020one}. Input devices such as VR controller\cite{zhang2018deep}, space mouse\cite{gubbi2020imitation}, visual odometry for 6-DoF position tracking using smartphones\cite{mandlekar2018roboturk}\cite{mandlekar2019scaling}, etc. have been used to gather expert demonstrations. What is common to all of these approaches is that some input device is used to enable human experts to teleoperate the robot. In this work, we deviate from that approach by having experts indirectly control the robot by writing Python programs.

Understanding natural language in the context of the visual scene of the robot has been addressed by several papers. In \cite{shridhar2018interactive}, a robot system to pick and place common objects is built where the object is inferred from the input image and grounded language expressions. Understanding instructions provided in spoken language with incomplete information based on the context of the input image and common sense reasoning is addressed in \cite{chen2020enabling}. The authors in \cite{liu2019clevr} propose a synthetic dataset for visual question answering to debug and understand weaknesses in different grounded natural language reasoning models. In \cite{bisk2016natural}, the Blocks dataset is proposed to evaluate grounded spatial reasoning capabilities of neural networks. Our work also has an emphasis on spatial reasoning, but we go beyond moving a single object. 

Unlike the above mentioned works, the Learning from Play (LfP) approach in \cite{lynch2020grounding} is goal-based imitation learning with the neural network directly controlling the actuators. Rather than conditioning on the target image, \cite{lynch2020grounding} replaces it with a latent vector derived from the natural language input. In this paper, we use the more traditional imitation learning approach and have experts translate natural language instructions into Python code. In Concept2Robot\cite{ShaoConcept2RobotLM}, a large dataset of human demonstrations (not teleoperated) is used to learn a reward function that is then used for training a policy network using reinforcement learning. In this work, we do not use reinforcement learning or a reward function and instead use the programs written by the expert in a fully supervised learning setting.

Much attention is devoted to object detection in the computer vision literature\cite{howard2017mobilenets}\cite{redmon2016you}\cite{ren2015faster}\cite{venkatesh2019one}\cite{venkatesh2020teaching}. Although end-to-end imitation learning does not use object detection, it is also possible to use a pipelined approach where object detection is one module. For example, in \cite{zeng2018robotic}, the pick-and-place task is performed by  picking up the object at a grasp point and then bringing it near the camera for classifying to which bin the object should be placed in. In this paper, we use a fully convolutional object detector inspired by \cite{redmon2016you} to detect the positions and sizes of all the objects in the scene.


The problem of answering queries in natural language using data from a table is addressed in \cite{herzig2020tapas}. There are broadly two approaches to this problem. One way is to approach this as a semantic parsing problem and to generate a logical form or a SQL query from the natural language input. The other way is to process the natural language instruction along with the contents of the table to directly predict the answer. The latter approach subsumes the process of running the query into the neural network itself. In this paper, we generate Python function blocks rather than SQL statements from natural language.

In \cite{barone2017parallel}, the authors propose generating code from documentation strings. In \cite{feng2020codebert}, a pre-trained model for  programming languages is proposed. A ``transpiler" that translates code from one language to another is proposed in \cite{lachaux2020unsupervised}. Although this paper also proposes generating program code from natural language, the end goal of controlling the robot is different. As a result, the evaluation metrics and baselines also differ. Moreover, our primary objective in this work is not to improve on code generation methods, but to show that generating code can outperform direct prediction of actuator commands.

\section{TASK DESCRIPTION}

We consider two different tasks where the task is specified using natural language.

\subsection{Arrange task}

\begin{figure}[!t]
    \centering
    \includegraphics[width=1.0\linewidth]{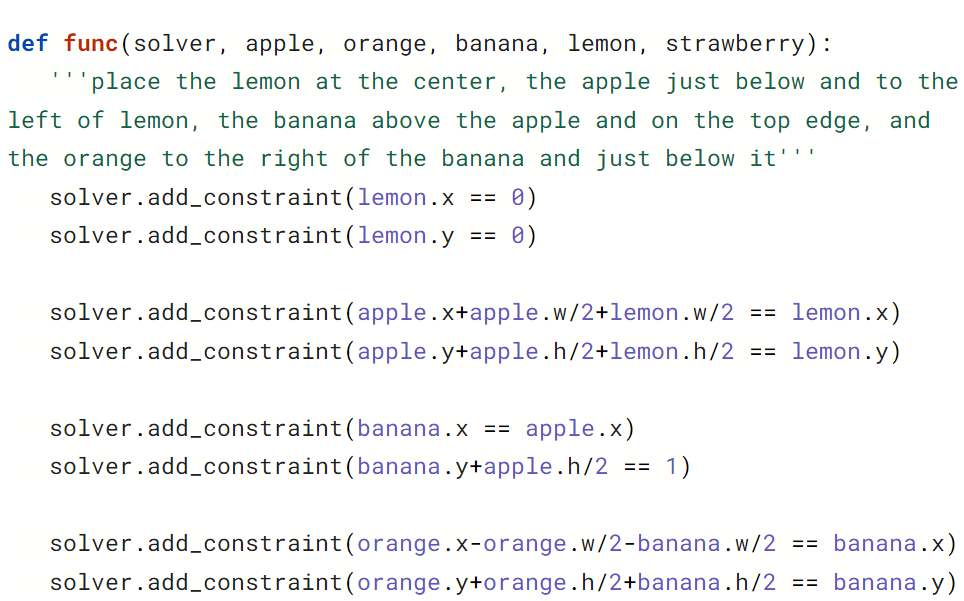}
    \caption{A sample function for the arrange task that takes the width and height of all the objects and determines the positions of the objects on the table as specified by the natural language instruction (which is shown in the docstring). The Cassowary constraint solver is used to determine the positions of the objects. Note that the extents of the table on which the objects are to be placed is normalized to be in the range $[-1, 1]$.}
    \label{fig:arrange_ex0}
\end{figure}

\begin{figure}[!t]
    \centering
    \includegraphics[width=1.0\linewidth]{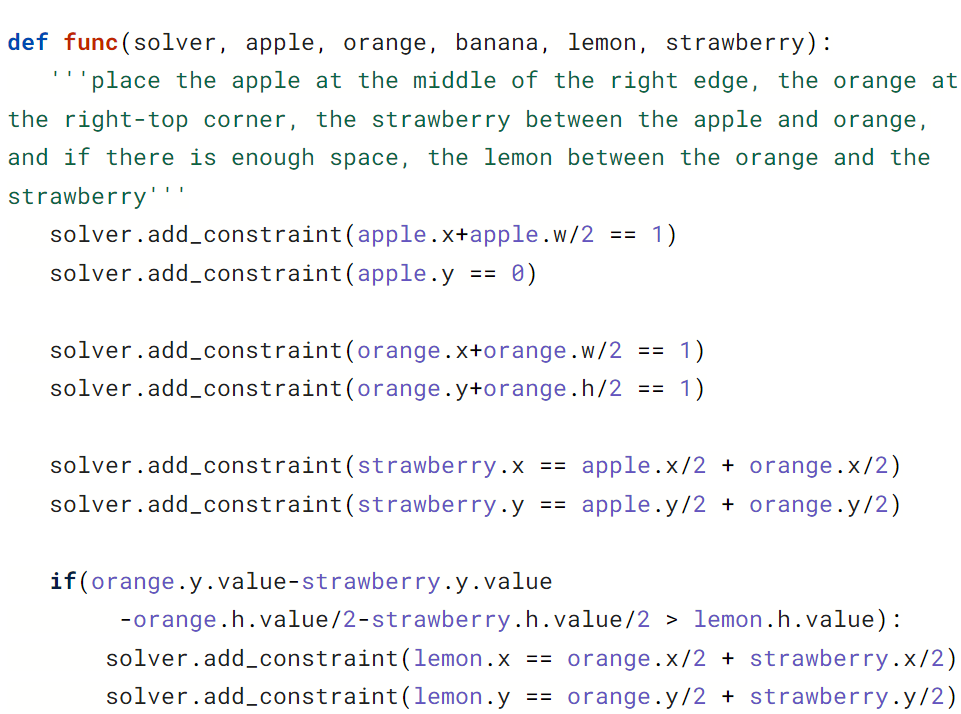}
    \caption{A sample function for arrange task that uses the output of the constraint solver before deciding to add additional constraints.}
    \label{fig:arrange_ex1}
\end{figure}

This task involves taking objects from a tray and placing them at different positions on the table. The instruction in natural language along with the width and height of all the objects are the inputs and the goal is to predict the positions of the objects on the table. The motion planning to pick up the object from the tray and place it at the specified location is performed separately (this is not learned).

Figures~\ref{fig:arrange_ex0} and \ref{fig:arrange_ex1} show sample programs that compute the positions of the objects for the given natural language instruction.  The program uses the Cassowary constraint solver (which uses the simplex method) to declaratively specify constraints for the positions of the objects. Note that it's not entirely declarative and the program can access the intermediate solution before declaring additional constraints (Fig.~\ref{fig:arrange_ex1}). After the program is executed, the positions of all the objects determined by the constraint solver are used to plan the pick-and-place motion of the robot arm.

\subsection{Manipulation task}

\begin{figure}[!t]
    \centering
    \includegraphics[width=1.0\linewidth]{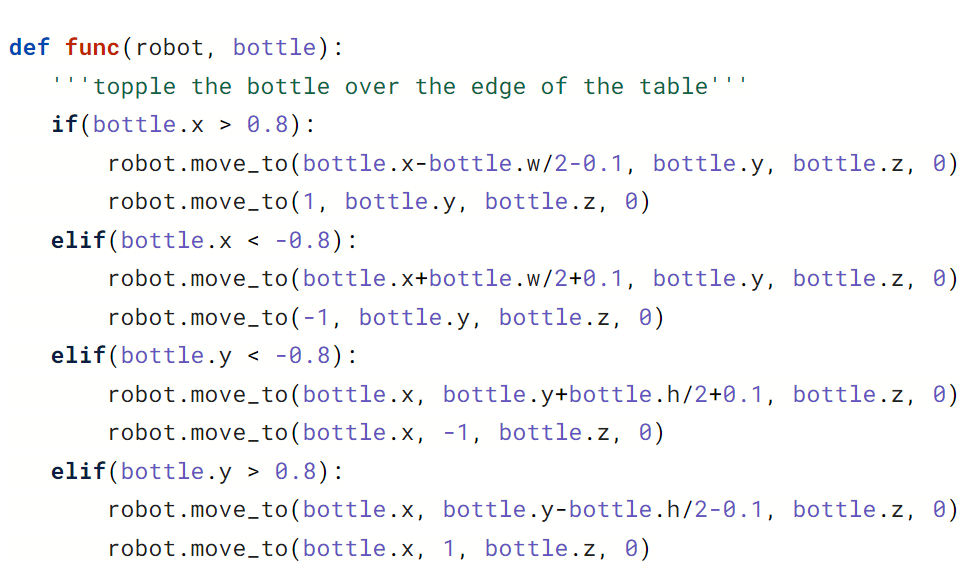}
    \caption{A sample function for the manipulation task that takes the positions and sizes of all the objects on the table and determines the sequence of robot actions to accomplish the goal  specified by the natural language instruction (which is shown in the docstring). The program can control the robot by specifying the end-effector position and the suction gripper state (on/off).}
    \label{fig:manipulation_ex0}
\end{figure}

\begin{figure}[!t]
    \centering
    \includegraphics[width=1.0\linewidth]{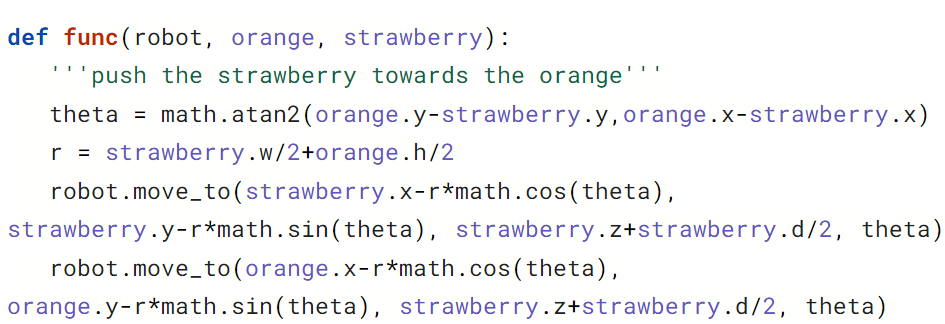}
    \caption{A sample function for manipulation task that uses trigonometric functions in the Python standard library to compute the trajectory of the end-effector of the robot.}
    \label{fig:manipulation_ex1}
\end{figure}

This task involves manipulating objects on the table as specified by the natural language instruction. Typical tasks involve reaching for an object, pushing an object somewhere, and picking-and-placing an object. To control the robot, the action space is (a) to move the end effector of the robot to the specified position (x, y, z, r), and (b) to control the suction gripper (on/off). The robot can be controlled by emitting a sequence of end effector poses and grip commands. An object detector makes available the positions and sizes of all the objects. The goal is to take the positions and sizes of all the objects on the table and to emit a sequence of end-effector positions and gripper on/off commands.

Figures~\ref{fig:manipulation_ex0} and \ref{fig:manipulation_ex1} show sample programs that control the robot to accomplish the task specified by the natural language instruction. Unlike the previous task, the objects are already on the table. Moreover, the program must not merely specify the desired state, but it must also directly control the robot to get to the desired state. So, the current positions of the objects are used to compute the appropriate actions.

\section{NETWORK ARCHITECTURE}\label{sec:arch}

\begin{figure*}[!t]
    \centering
    \includegraphics[width=0.85\linewidth]{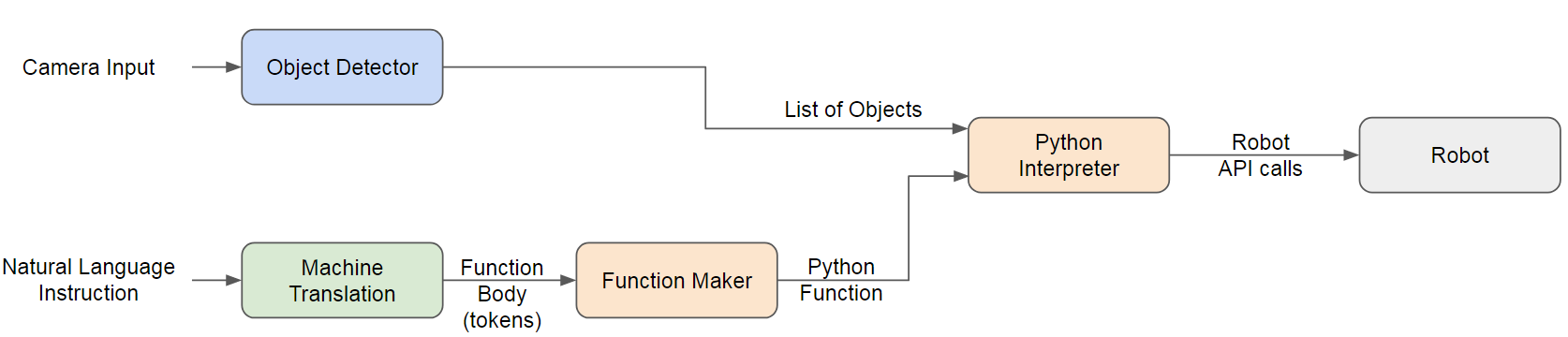}
    \caption{Illustration of the system architecture. The natural language instruction is translated into the tokens of a Python program using neural machine translation. These tokens are assembled into an anonymous Python function. The list of objects in the scene from the object detector are passed as arguments to this function. Object classes that are not present in the scene are set to ``None" when calling the generated function and will cause an exception if the generated function attempts to read their properties. When the function is executed, it communicates with the robot via a simple API consisting of ``move" and ``grip" commands.}
    \label{fig:sys_arch}
\end{figure*}

\begin{figure*}[!t]
      \centering
      \includegraphics[width=0.75\linewidth]{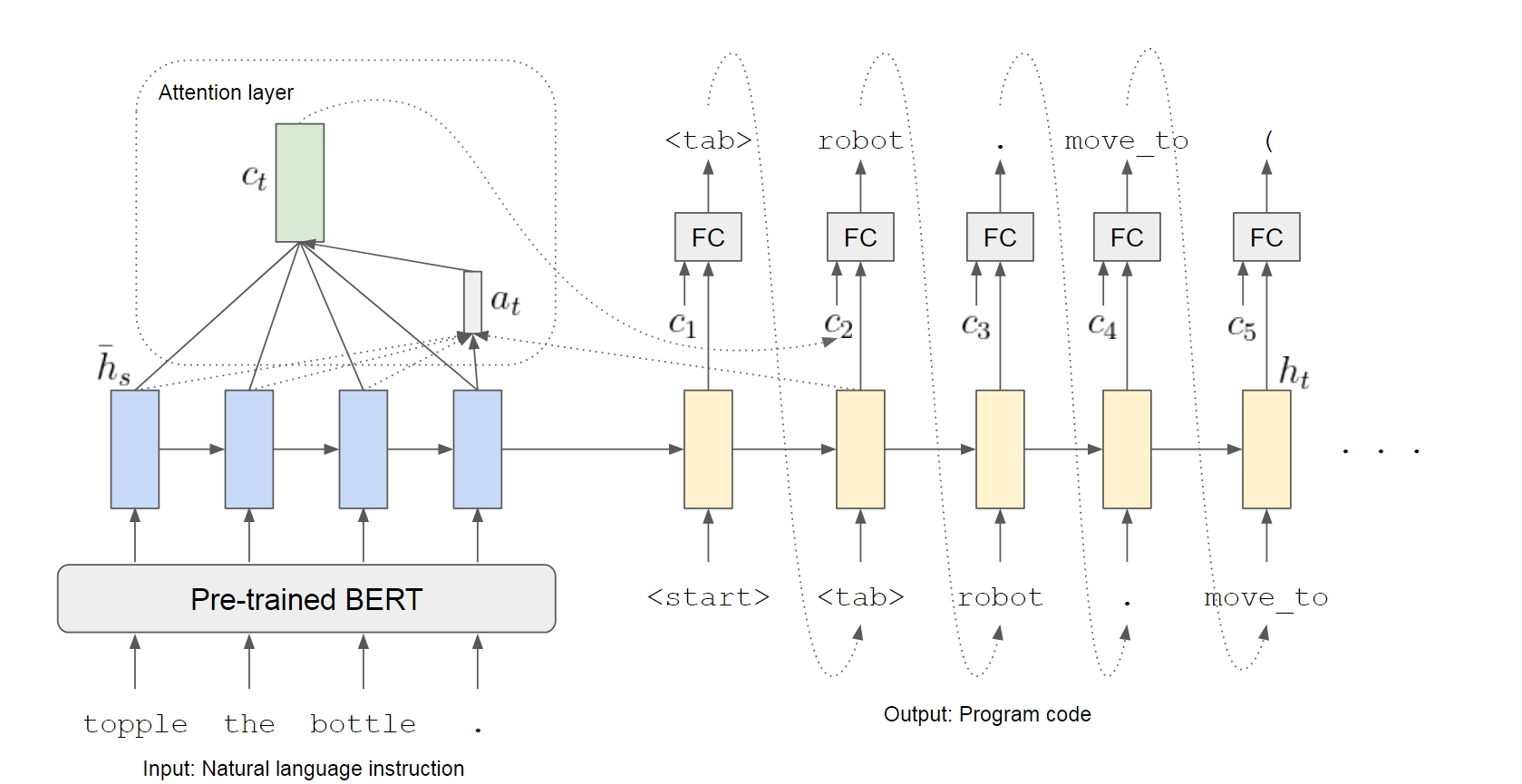}
      \caption{The proposed neural machine translation architecture. The recurrent cells used are a single layer of LSTM cells with hidden state dimension of 1024. In the decoder, the context vector $c_t$ is concatenated with $h_t$ and passed through fully connected layers (FC1024-ReLU-FC100-Softmax) to predict the target sequence.}
      \label{fig:arch}
\end{figure*}

The overall system architecture is depicted in Fig.~\ref{fig:sys_arch} with an example in Fig.~\ref{fig:hero2}. The image from the camera and the instruction in natural language are the inputs, and a Python program that drives the robot is the output. The image from the camera is processed by an object detector to obtain a list of objects in the scene. For each object in the scene, the object detector provides its position and size. We use a single shot detector\cite{liu2016ssd} fine-tuned to detect the objects we use, but any object detector can be used without affecting the rest of the system. The natural language instruction is processed by a neural machine translation model described in the following paragraphs to translate the instruction into tokens of a Python function body (Fig.~\ref{fig:arch}). The Python function is constructed by listing all the detected objects as arguments to the function and concatenating the generated tokens to form the function body. When executed, this function drives the robot to perform the task specified by the natural language instruction. Note that the objects that can be detected are fixed, and the translation model can only generate tokens corresponding to the known objects (present in the training set). However, the translation model can potentially generate arbitrarily complex programs. 


We use an LSTM based neural machine translation model with attention\cite{luong2015effective}. Unlike most language vision models, our translation model does not take the image observation as an input. Rather, the program generated by the network accesses the attributes of the objects detected and controls the robot based on that.

The input natural language instruction is tokenized, and the embeddings for the tokens are obtained using a pre-trained BERT model\cite{devlin2018bert}. Note that the BERT layers are frozen and remain unchanged during training. The input sequence embeddings are processed by an encoder LSTM with hidden states $\bar{h}_s$. After all the input tokens are processed, a decoder LSTM predicts the target sequence that is used to construct the Python function body.

At each step of the decoder, the decoder state $h_t$ is used to attend to the input states and infer the context vector $c_t$ that is used to predict the output $y_t$.

The variable length alignment vector $a_t$ of size equal to the number of steps in the input sequence is obtained by comparing the decoder hidden state $h_t$ with each of encoder hidden states $\bar{h}_s$:

\begin{equation}
    \hat{a}_t(s) = h_t^T W_a \bar{h}_s
\end{equation}

\begin{equation}
    a_t(s) = \frac{\exp(\hat{a}_t(s))}{\sum_{s'}\exp(\hat{a}_t(s'))}
\end{equation}

The context vector $c_t$ is computed as the weighted average of the hidden states of the encoder $\bar{h}_s$:

\begin{equation}
    c_t = \sum_{s}a_t(s)\bar{h}_s
\end{equation}

The context vector $c_t$ and the decoder state $h_t$ are concatenated and passed through fully connected layers to predict the target sequence token $y_t$.

\section{RESULTS}

There are several parts in the proposed system, and each of them can cause the robot to fail in performing the specified task. The object detector might be inaccurate, the generated program might be incorrect, or the end-effector might not successfully pick an object (for example, suction fails to lift an object). Although we are ultimately interested in whether the robot can successfully complete the task specified, it is useful to isolate the proposed machine translation model and analyze it independently. For this, we assume that the object detector is perfectly accurate and the robot is always successful in moving and picking objects. Under this ideal scenario, we first evaluate the proposed approach. Subsequently, we discuss the performance of the complete system on a real robot arm.

The proposed method is evaluated using a desktop workstation with Intel Core~i7-4790K processor, 32~GB RAM, and nVidia RTX~2080Ti GPU. The models are trained using Tensorflow~1.14. A USB webcam (Logitech~C310) attached to the workstation is used to capture images. The single shot object detector (from Tensorflow hub) finds the objects in the scene, and the translation model generates the program from the natural language instruction. We wrote a thin custom wrapper around the ``pydobot" package to provide a simple Python API that the generated program uses to control the Dobot Magician robot arm which is connected to the workstation via USB. The pre-trained BERT model, which we use in our translation moel, is obtained from the HuggingFace Transformers library\cite{wolf-etal-2020-transformers}. 

\subsection{Datasets}

\subsubsection{Arrange Dataset} The arrange task involves arranging objects on the table as specified by the instruction in natural language. For this task, we have collected the arrange dataset, a parallel corpus of instructions in English and Python functions. The function takes the object sizes as arguments and sets the position of the objects as indicated in the instruction. Some examples are shown in Figs.~\ref{fig:arrange_ex0} and \ref{fig:arrange_ex1}. Note that in addition to the object sizes, the function is also given the Cassowary linear constraint solver\footnote{The Cassowary  algorithm is used by Apple UIKit to place UI elements in GUIs} to specify the positions of objects as constraints to be solved. The arrange dataset has training / development / test split of 102 / 11 / 11 samples. These samples are augmented by randomly changing the object(s) simultaneously in both the instruction text and the program.

We also execute each ground truth program in the corpus for 20 different random initializations of the sizes of the objects to obtain the positions of the objects given those sizes. This secondary dataset is used for fair comparison with baseline models that directly predict the positions of the objects given the instruction and sizes of the objects.

\subsubsection{Manipulation Dataset} This task involves manipulating objects already present on the table as specified by the instruction in natural language. Typical manipulation tasks in this dataset are reaching for an object, pushing an object somewhere, and picking-and-placing an object. For this task, we have collected the manipulation dataset, a parallel corpus of instructions in English and Python functions. The function takes the positions and sizes of all the objects on the table and controls the robot through an API that allows it to specify a sequence of end-effector poses and gripper states (on/off). A few examples are shown in Figs.~\ref{fig:manipulation_ex0} and \ref{fig:manipulation_ex1}. The manipulation dataset has training / development / test split of 122 / 12 / 12 samples. These are augmented by randomly changing the object(s) simultaneously in both the instruction text and the program.

For each sample in the manipulation corpus, the ground truth Python program is executed for 20 random initializations of the positions and sizes of the objects on the table and with a mock robot that records the sequence of end-effector positions and gripper state changes. This is used for fair comparison with baseline models that directly predict the sequence of end-effector poses given the instruction text and the sizes and positions of the objects.

\begin{figure}[!t]
    \centering
    \includegraphics[width=1.0\linewidth]{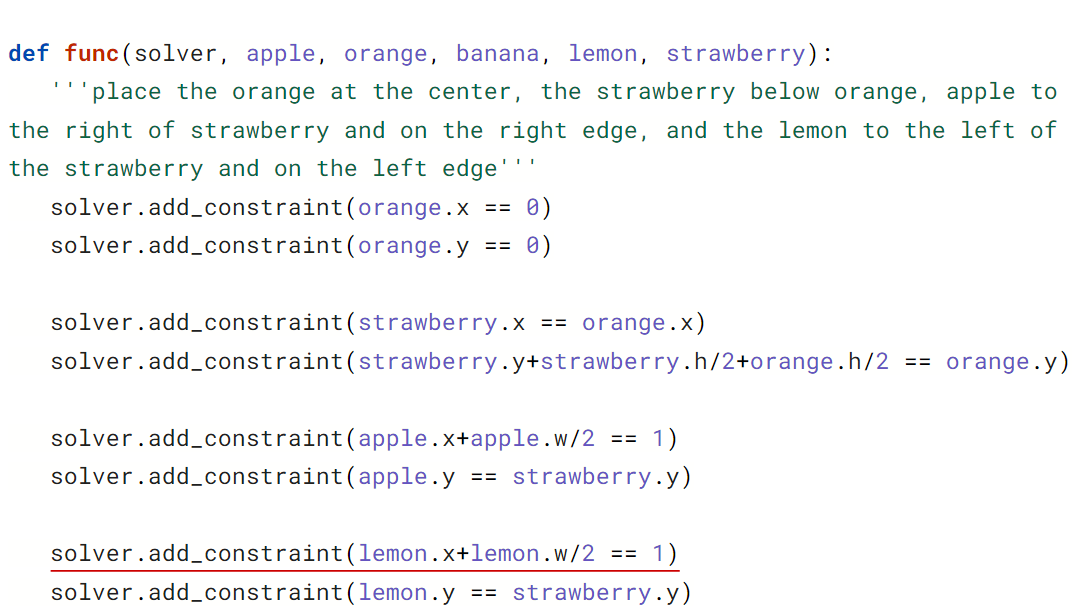}
    \caption{Sample generated program (incorrect) from the test set for the arrange task. The input instruction is in the docstring. The underlined code is incorrect. The neural network seems to have overfit for instructions with multiple phrases, and the generated code resembles a sample in the training set.}
    \label{fig:res-arrange0}
\end{figure}

\begin{figure}[!t]
    \centering
    \includegraphics[width=1.0\linewidth]{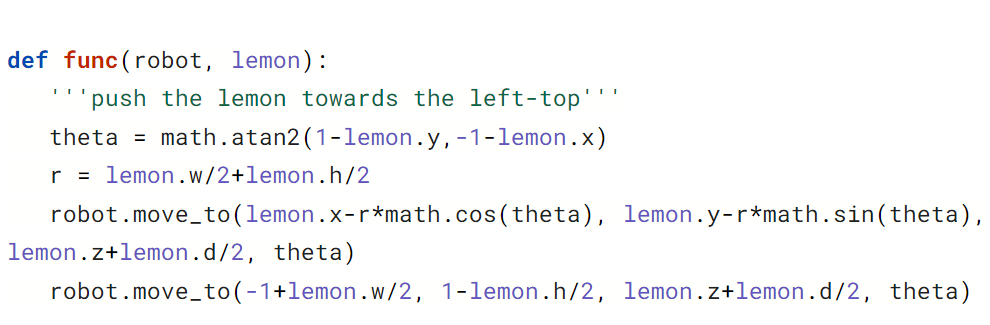}
    \caption{Sample generated program (correct) from the test set for the manipulation task. The input instruction is in the docstring.  The program moves the end-effector to push the object to the intended location.}
    \label{fig:res-misc0}
\end{figure}

\begin{figure}[!t]
    \centering
    \includegraphics[width=1.0\linewidth]{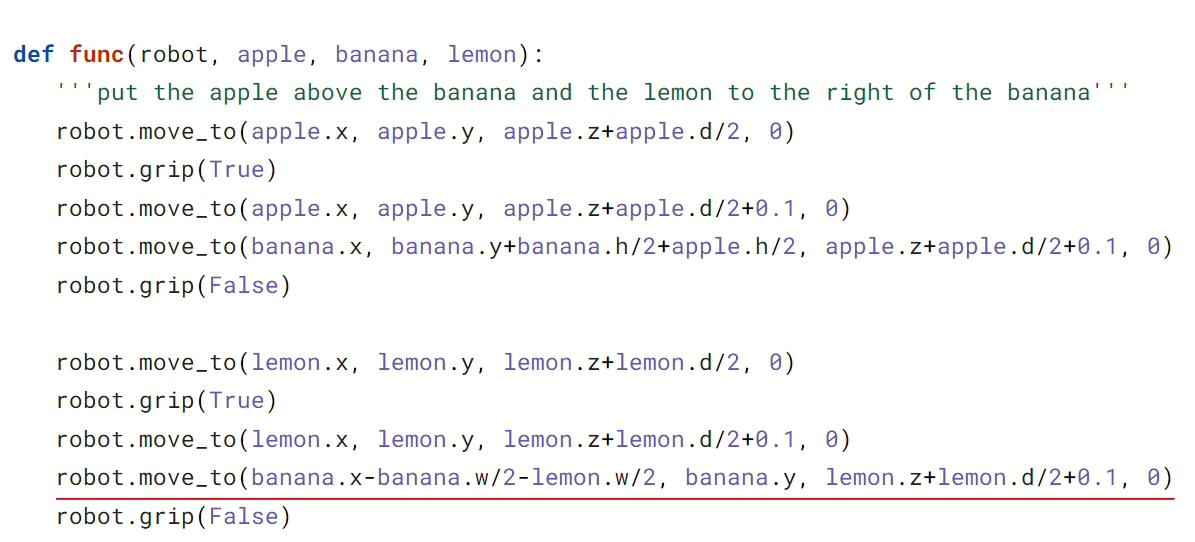}
    \caption{Sample generated program (incorrect) from the test set for the manipulation task. The input instruction is in the docstring. The underlined code is incorrect. The network seems to have overfit since the incorrect generated program resembles a sample in the training set.}
    \label{fig:res-misc1}
\end{figure}

\begin{figure}[!t]
    \centering
    \includegraphics[width=1.0\linewidth]{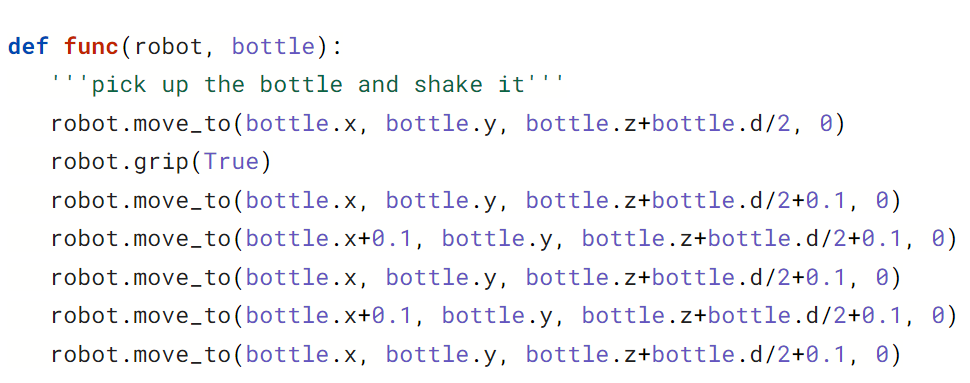}
    \caption{Sample generated program (correct) from the test set for the manipulation task. The input instruction is in the docstring. The program successfully controls the robot to perform the task.}
    \label{fig:res-misc2}
\end{figure}

\subsection{Baselines}\label{sec:baselines}

For the arrange dataset, we use LSTM+FC layers\cite{bisk2016natural} as the baseline. The LSTM encodes the instruction text into a fixed size vector. This is concatenated with the sizes of all the objects and passed through several fully connected layers to directly predict the positions of all the objects.

For the manipulation dataset, we use an encoder LSTM to encode the instruction and a decoder LSTM that, at every timestep, concatenates the decoder state and the attention context vector at that timestep along with the positions and sizes of all the objects on the table, and passes this concatenated vector through fully connected layers to predict the end-effector pose and grip state\cite{rahmatizadeh2018virtual}. 

\subsection{Evaluation Metric}

Evaluating a machine translation model that generates programs is challenging. Traditionally used metrics for translation models such as the BLEU score that measures the similarity between the predicted and ground truth tokens do not work well for measuring similarity of program code. The slightest change (a `+' to a `-') might give a high BLEU score but result in catastrophic failure, whereas dis-similar looking programs might actually be expressing the same logic\cite{lachaux2020unsupervised}. We could simply check to see if the generated program is syntactically valid. But, this overestimates performance since many generated programs that are syntactically valid might still not accomplish the specified task. On the other hand, if we check to see if the generated program exactly matches the ground truth, we would underestimate performance since the same logic can be expressed in myriad ways. So, the best way to evaluate the generated code is to execute it and to check what it does\cite{lachaux2020unsupervised}. Thus, we choose an extrinsic metric to evaluate the goodness of the generated program.

We use accuracy as the evaluation metric. Each of the predicted programs are executed 20 times with randomized object positions and sizes. Our custom robot API allows us to capture the actuator commands generated by the Python program driving the robot. During training and validation, the robot API calls (such as ``move" and ``grip") along with the arguments are merely recorded and not sent to the robot. Thus, we can execute both the generated program and the ground truth program and compare the resulting end-effector trajectories. For the arrange dataset, we treat the prediction to be ``correct" if the absolute difference between the predicted position and ground truth position is less than 10\% of the width of the table (on both x and y axes). Although the natural language instruction might admit multiple solutions (for example, ``place the apple to the left of the orange" is under-specified), all the labelled data have a canonical, unambiguous target position, which eliminates any difficulty in measuring accuracy. For the manipulation dataset, the prediction is considered accurate if the absolute difference between the predicted trajectory and the ground truth trajectory is less than 10\% of the width of the table at every timestep. This is merely an easy-to-evaluate proxy for whether the robot is truly accomplishing the task in the instruction. A more thorough evaluation that properly tests whether the task specified was performed successfully is conducted on a few samples with a real robot arm (Section~\ref{sec:real-robot}).

\subsection{Discussion of Results}

\begin{figure}[!t]
    \centering
    \includegraphics[width=1.0\linewidth]{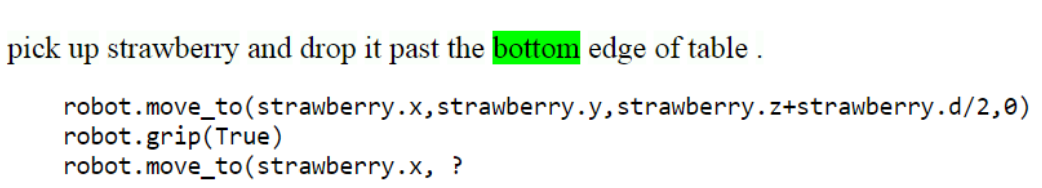}
    \caption{Visualization of attention over the natural language instruction when predicting the token denoted by ``?". When predicting the y-coordinate of the move function call, the attention layer is focusing on the ``bottom" edge of the table to emit the ``-1" token. }
    \label{fig:attn-vis0}
\end{figure}

\begin{figure}[!t]
    \centering
    \includegraphics[width=1.0\linewidth]{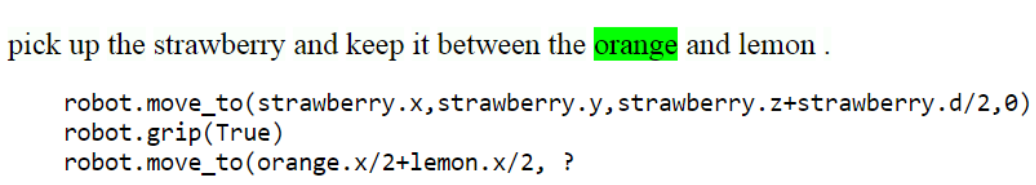}
    \caption{Visualization of attention over the natural language instruction when predicting the token denoted by ``?". When predicting the y-coordinate of the move function call, the attention layer is focusing on the ``orange" in the input instruction to emit the ``orange" token.}
    \label{fig:attn-vis1}
\end{figure}

\begin{table}[!t]
\caption{Comparison of the performance of baselines (Section~\ref{sec:baselines}) and the proposed model (Section~\ref{sec:arch})}
\label{tab:results}
\begin{center}
\begin{tabular}{lcc}
\hline
Model & Arrange Task & Manipulation Task\\
\hline
Baseline (\cite{bisk2016natural} / \cite{rahmatizadeh2018virtual}) & 14.2\% & 9\%\\
\hline
Proposed Seq2Seq model & 80.8\% & 93.2\%\\
\hline
\end{tabular}
\end{center}
\end{table}

Table~\ref{tab:results} compares the results of the proposed method with the baselines. All the architectures are trained using the Adam optimizer (learning rate 1e-3) for 10 epochs with batch size of 16. For both tasks, the proposed method of generating a Python program and then executing that program outperforms the baselines which directly regress the object positions (arrange dataset) or end-effector poses (manipulation dataset). The percentage of generated programs that were malformed (due to syntax errors) and could not be run were 0.6\% and 2.81\% in the arrange dataset and manipulation dataset respectively. The baseline models perform significantly worse on our dataset than other datasets\cite{bisk2016natural} and also compare poorly with the translation model because we are attempting to train, with limited training data, a neural network to learn to solve linear constraints (Fig.~\ref{fig:arrange_ex1}). However, in the proposed method, the constraint solver is presented as a readily available tool which the translation model only needs to learn how to employ. In the manipulation task, the same instruction text can result in very different robot end-effector trajectories depending on the positions of the objects (for example, ``push the bottle off the edge of the table"). The program captures all possible trajectories concisely and also the switch-over points when the trajectory changes because the location of the relevant object has changed (Fig.~\ref{fig:manipulation_ex1}). In contrast, the teleoperated expert demonstrations capture the trajectory only for the position of the object in that sample and offer no clue as to when a different trajectory is suitable for the same instruction text.

Figures~\ref{fig:res-arrange0}-\ref{fig:res-misc2} show a few programs generated from the test set. Figures~\ref{fig:attn-vis0} and \ref{fig:attn-vis1} show the attention weights for different tokens of the input instruction text when predicting a particular output token. We see that the attention mechanism is focusing on the relevant part of the instruction when predicting the program. 

We have also experimented with replacing words in the instruction text with synonyms. We found that replacing ``put" with ``keep", ``place", and ``put down" always resulted in correct predictions. Likewise, we found that removing the word ``the" does not change the output. Similarly, replacing ``right-top corner" with only ``right-top" or ``top right" results in no changes to predicted sequence. However, substituting the words for objects, such as replacing ``bottle" with ``flask" or ``pitcher" and ``cup" with ``chalice", caused incorrect predictions. Also, deleting the object from the instruction text resulted in incorrectly generated programs that had syntax errors. Although our datasets are small, these findings suggest that it is worth investigating the proposed method of using neural machine translation for code generation with larger datasets. Similar recent efforts using GPT-3\cite{brown2020language} to generate code also bolster the case for further investigation.

We also found that the generalization worsens as the number of phrases in the input sequence increases (Figs.~\ref{fig:res-arrange0} and \ref{fig:res-misc1}). There are only a few samples in the training set with 4 phrases (such as ``place the orange at the bottom-right, the apple at the top-right, banana at the center, and the lemon to the right of the apple"). The model overfits on such long phrases and gives incorrect predictions that resemble the training data. However, if the input instruction is split at the commas into multiple short phrases, the model correctly predicts the positions for each of the phrases. But, this is not a viable solution because there are many instructions where such a split is not possible since the latter phrases refer to objects in the former (for example, ``place the apple at the center, the orange at the top-right, and the banana in between them").

\subsection{Demonstration on the Robot Arm}
\label{sec:real-robot}

We demonstrate the complete pipeline with a Dobot Magician (Fig.~\ref{fig:hero}). Common objects such as fruits, cups, magnets, etc. are used. An object detector\cite{liu2016ssd} is trained to detect the position and size of these objects, but the depth (tallness from the table surface) of the object is measured beforehand and hard coded. The camera feed from an overhead camera is passed through the object detector whose output is passed as arguments to the Python function generated by the proposed method from the natural language instruction, and the function is executed. Out of 25 trials, 19 were successful with the robot accomplishing the task. All the failures were due to inaccuracies in the object detector or the suction gripper failing to pick up the object (the few longer instructions which systematically caused translation errors were not present in the small sample of instructions tested with the real robot).
A video of the robot in operation is available at: \url{https://youtu.be/usCvsDIgWOM}

\section{CONCLUSIONS}

We find that programs are rich representations of the expert demonstrations and are beneficial for learning to control robots. Moreover, the predicted programs are interpretable and easier to analyse than end-to-end neural networks that directly predict robot actions. Although this approach is necessarily constrained to those problems for which the solution can easily be expressed as a program, the proposed approach may find use in augmenting teach pendants for industrial robots to generate programs based on verbal instructions. The proposed method of generating programs is promising, so it could be worth investigating if performance can be improved by pre-training the program generator on a large corpus of source code. The proposed approach could also be useful in enabling an easily interpretable conversational system where the robot can ask clarifying questions.


\addtolength{\textheight}{-2cm}   




\section*{ACKNOWLEDGMENT}

We thank Mohammed Rizvi for his suggestions and the Robert Bosch Center for Cyber-Physical Systems for funding support.


\bibliographystyle{IEEEtran}
\bibliography{IEEEabrv,mybibfile}

\end{document}